%% file: conf.tex
\documentclass[conference]{IEEEtran}

\input{config.tex}

\input{macros.tex}

\begin{document}

\title{Feature learning in feature--sample networks using multi-objective optimization}

\author{
  \IEEEauthorblockN{Filipe Alves Neto Verri}
  \IEEEauthorblockA{
      \small
      Institute of Mathematical and Computer Sciences \\
      University of São Paulo -- São Carlos, SP, Brazil \\
      \vspace{-0.19in}
      E-mail: filipeneto@usp.br
  }
  \and
  \IEEEauthorblockN{Renato Tinós, Liang Zhao}
  \IEEEauthorblockA{
      \small
      Department of Computing and Mathematics, FFCLRP \\
      University of São Paulo -- Ribeirão Preto, SP, Brazil \\
      E-mail: rtinos@ffclrp.usp.br, zhao@usp.br
  }
}

\maketitle

\input{abstract.tex}
\input{intro.tex}
\input{problem.tex}
\input{optimization.tex}
\input{result.tex}
\input{conclusion.tex}

\bibliographystyle{IEEEtran}
\bibliography{IEEEabrv,references}

\end{document}

%% file: config.tex
\usepackage[utf8]{inputenc}
\usepackage[english]{babel}

\usepackage{cite}

\usepackage[cmex10]{amsmath}
\usepackage{amsfonts}
\usepackage{amssymb}
\usepackage{mathtools}
\usepackage{algorithm}
\usepackage{algpseudocode}

\usepackage[pdftex]{graphicx}
\graphicspath{{./images/}}

\usepackage{booktabs}
\usepackage{paralist}
\usepackage{color}

\usepackage{hyperref}
\usepackage{cleveref}
\crefname{section}{Section}{Sections}
\crefname{algorithm}{Algorithm}{Algorithms}
\crefname{equation}{Equation}{Equations}
\crefname{figure}{Figure}{Figures}
\crefname{table}{Table}{Tables}
\crefname{theorem}{Theorem}{Theorems}
\crefname{lemma}{Lemma}{Lemmas}

%% file: macros.tex
\newcommand{\FSNetwork}{\mathcal{G}}
\newcommand{\Sample}[1][i]{\vec{x}_{#1}}
\newcommand{\Feature}[1][ij]{x_{#1}}
\newcommand{\VertexSet}{\mathcal{V}}
\newcommand{\EdgeSet}{\mathcal{E}}

\newcommand{\card}[1]{\left|#1\right|}

\newcommand{\pipe}{\left\vert\right.}

\newcommand{\Disproportion}{\Delta}
\newcommand{\Degree}[1][i]{k_{#1}}

%% file: abstract.tex

\begin{abstract}

  Data and knowledge representation are fundamental concepts in machine
  learning.  The quality of the representation impacts the performance of the
  learning model directly.
  Feature learning transforms or enhances raw data to structures that are
  effectively exploited by those models.
  In recent years, several works have been using complex networks for
  data representation and analysis.  However, no feature learning method has
  been proposed for such category of techniques.
  Here, we present an unsupervised feature learning mechanism that works on
  datasets with binary features.
  First, the dataset is mapped into a feature--sample network.  Then, a
  multi-objective optimization process selects a set of new vertices to produce
  an enhanced version of the network.  The new features depend on a nonlinear
  function of a combination of preexisting features.  Effectively, the process
  projects the input data into a higher-dimensional space.
  To solve the optimization problem, we design two metaheuristics based on the
  lexicographic genetic algorithm and the improved strength Pareto evolutionary
  algorithm (SPEA2).
  We show that the enhanced network contains more information and can be
  exploited to improve the performance of machine learning methods.
  The advantages and disadvantages of each optimization strategy are discussed.

\end{abstract}


%% file: intro.tex

\section{Introduction}


A good representation of the encoded knowledge in a machine
learning model is fundamental to its success.  Several data structures have been
used for this purpose, for instance, matrices of weights, trees, and
graphs~\cite{Bishop2006,Zhu2009}.  Sometimes, learning models lack a data
structure to store knowledge, storing the input as is \cite{Cover1965}.

In recent years,
several works have been using \emph{complex networks} for data representation and
analysis~\cite{Thiago2012e,SilvaZhao2016,Verri2016b}.
Complex networks are graphs with a nontrivial topology that represent the
interactions of a dynamical system~\cite{Newman2010}.  Advances in the science
of complex systems bring several tools to understand such systems.

In~\cite{Verri2016}, we describe how to map a dataset with binary features
into a bipartite complex-network.  Such network is called \emph{feature--sample network}.
Using that representation, we solve a semi-supervised learning task called positive-unlabeled (PU)
learning~\cite{Munoz-Mari2010}.

When dealing with machine learning problems, we often need to pre-process the
input data.  \emph{Feature learning} transforms or enhances raw data to
structures that are effectively exploited by learning models.  Autoencoders
and manifold learning are examples of feature learning methods~\cite{Bengio2013}.

In this paper, we propose a feature learning process to add information in
feature--sample networks.  In summary, we include to the network a limited
number of new vertices based on a non-linear function of the preexisting ones.
The set of new vertices is determined by a multi-objective objective problem, in which
the goal is to maximize the number of features while maintaining some properties of the
original data.  Two multi-objective approaches are designed: a lexicographic
genetic algorithm (LGA) and an implementation of the improved strength Pareto
evolutionary algorithm (SPEA2).

We show that enhanced feature--sample networks improve the performance of
learning methods in the major machine learning paradigms: supervised and unsupervised learning.
We also expose the pros and cons of each optimization approach.

The rest of this paper is organized as follows.  \Cref{sec:problem,sec:methods}
describe how to enhance feature--sample networks as an optimization problem.  In
\cref{sec:results}, computer simulations illustrate the optimization process and
assess the performance improvements in machine learning tasks.
Finally, we conclude this paper in \cref{sec:conclusion}.

%% file: problem.tex


\section{Enhanced feature--sample networks}
\label{sec:problem}

In this section, we describe how we enhance a feature--sample network by
adding
to the network a constrained number of new features. Connections between the
samples and each new feature depend on a nonlinear function of a combination of
preexisting features.  The chosen set of new features is the result of an
optimization process that not only maximizes the number of new features but also
distributes them along the samples.  This process is similar to the projection of
the associated data into a higher-dimensional space preserving some
important characteristics of the data.

In the following subsections, we first review the feature--sample networks, then
describe the creation of new features.  Moreover, we elaborate an optimization
problem to enhance feature--sample networks.

\subsection{Feature--sample network review}

Assume as input the dataset $\mathcal{B} = \{\Sample[1], \ldots, \Sample[N]\}$ whose
members are binary feature vectors $\Sample[i] = [\Feature[i1], \dots,
\Feature[iD]] \in \{0, 1\}^D$.  The feature vectors are sparse, that is, the number of
elements with value 1 is much lower than the dimension $D$.

The \emph{feature--sample network} $\FSNetwork$
is the bipartite complex-network whose edges connect samples
and features of the dataset $\mathcal{B}$.  A simple, unweighted, undirected graph
$(\VertexSet, \EdgeSet)$ represents such network.  The vertex set $\VertexSet$
is $\{v_1, \dots, v_N, v_{N+1}, \dots, v_{N+D}\}$ and an edge exists between
\emph{sample} $v_i$ and \emph{feature} $v_{N+j}$ if $\Feature[ij] = 1$.

\subsection{And-features definition}
\label{sub:and-feature}

According to the \emph{Cover's Theorem} \cite{Cover1965}, given a not-densely
populated space of a classification problem, the chance of it being linearly
separable is increased as one cast it in a high-dimensional space nonlinearly.

Since we assume the input feature--sample network is sparse, we can synthesize
features nonlinearly to exploit the properties of this theorem.  One way to
produce new features is using the \emph{and} operator, which is a nonlinear
Boolean function.


We call \emph{and-feature} the feature $v$ that links to all samples
connected to a given set of two or more preexisting features.

Given a feature--sample network $\FSNetwork$ with $N$ samples and $D$ features,
we can produce an and-feature $v$ for each combination $\mathcal{W}$ of
$q$ features such that $\card{\mathcal{W}} \geq 2$ and $\mathcal{W} \subseteq
\{v_{N+1}, \dots, v_{N+D}\}$.  We call $q$ the order of the and-feature $v$.
Thus, the number of possible and-features is
\begin{equation*}
  \sum_{q = 2}^D{
    D \choose q
  } =
  \sum_{q = 2}^D{
    \frac{D!}{q!\left(D-q\right)!}
  } =
  2^D - D - 1\text{.}
\end{equation*}

In the rest of this paper, we index every possible and-feature using the
parameter $\vec{a} = [a_1$, \dots, $a_D] \in \{0, 1\}^D$ such that $\sum_j{a_j}
\geq 2$.  Each element $a_j$ indicates a feature in the network.  The feature
$v_{N+j}$ is part of the combination if, and only if, $a_j=1$.  Thus, the set
$\mathcal{W}$ is $\{v_{N+j} \pipe a_j = 1\}$.

Using this notation, we say that the and-feature $v(\vec{a})$ connects
to each sample $v_i$ if, and only if,
$
  \left(\lnot a_1 \lor x_{i1}\right) \land
  \dots \land
  \left(\lnot a_D \lor x_{iD}\right) = 1
$
holds.

From this discussion, we realize that enumerating every combination has
exponential cost.  Moreover, once the network is sparse, we expect that
many of the and-features have no connections.

\subsection{Optimization problem definition}

The problem of enhancing the network can be viewed as a optimization problem.
Given an input feature--sample network $\FSNetwork$ with $N$
samples and  $D$ features, we denote $\FSNetwork(\mathcal{Y})$ the enhanced
network from the original $\FSNetwork$ by adding every and-feature $v \in
\mathcal{Y}$.  The number of features of the enhanced network, excluding the
and-features that do not connect, is given by $D(\mathcal{Y})$.  Thus,
$\FSNetwork = \FSNetwork(\varnothing)$ and $D = D(\varnothing)$.

Let $M_{\text{max}}$ be the maximum number allowed of generated features, we
want to
\begin{equation*}
  \begin{aligned}
    & \underset{\mathcal{Y}}{\text{minimize}} & &
      D(\varnothing) - D(\mathcal{Y}) \\
    & \text{subject to} & &
      D(\mathcal{Y}) - D(\varnothing) \leq M_{\text{max}}\text{.}
  \end{aligned}
\end{equation*}

\newcommand{\FSbefore}{\FSNetwork}
\newcommand{\FSafter}{\FSNetwork'}

The disadvantage of this approach is that the and-features might not be well
distributed.  Thus, while some samples may have few new feature, others may
have many.  To overcome this
limitation, we introduce the \emph{disproportion} $\Disproportion(\FSbefore,
\FSafter) \in [0, \infty)$ between the network $\FSbefore$ and its enhanced version
$\FSafter$.

The disproportion is zero if the number of new connections in each sample is
proportional to its initial sparsity.  In this way, while the sparsity of each
sample might change, we keep the same shape of the degree distribution of the
samples.

Let $k_i$ be the degree of the sample $v_i$, the disproportion between two
networks is
\begin{equation*}
  \Disproportion(\FSbefore, \FSafter) = \mathrm{sd}\!\left(
    \frac{\Degree[1](\FSafter) - \Degree[1](\FSbefore)}{\Degree[1](\FSbefore)},
    \dots,
    \frac{\Degree[N](\FSafter) - \Degree[N](\FSbefore)}{\Degree[N](\FSbefore)}
  \right)\text{,}
\end{equation*}
where $\mathrm{sd}$ is the standard deviation of the arguments.

Using the disproportion in our optimization problem, the goal is to
\begin{equation*}
  \begin{aligned}
    & \underset{\mathcal{Y}}{\text{minimize}} & &
      D(\varnothing) - D(\mathcal{Y}),~%
      \Disproportion\big(\FSNetwork(\varnothing), \FSNetwork(\mathcal{Y})\big) \\
    & \text{subject to} & &
      D(\mathcal{Y}) - D(\varnothing) \leq M_{\mathrm{max}}\text{.}
  \end{aligned}
\end{equation*}


%% file: optimization.tex

\section{Methods}
\label{sec:methods}

In this section, we study the multi-objective problem stated in the previous
section and describe how we solve it.

\subsection{Problem study}

In \cref{sub:and-feature}, we see that the number of possible and-features
scales
exponentially in the number of features $D$.  As a consequence, storing every
possible and-feature is not feasible.

Furthermore, the number of candidate solutions is also exponential in the number
of possible new features.  Precisely, there are at the most
\begin{equation*}
  \sum_{m=1}^{2^D - D - 1}{
    {2^D - D - 1} \choose m
  } =
  \sum_{m=1}^{2^D - D - 1}{
    \frac{\left(2^D - D - 1\right)!}{m! \left(2^D - D - m - 1\right)!}
  }
\end{equation*}
solutions $\mathcal{Y}$ to explore.  We can not limit the size of the set
$\mathcal{Y}$ by $M_{\text{max}}$ since many and-features may have no
connection.

The three common approaches for solving multi-objective optimization problems
are weighted-formula, lexicographic, and Pareto~\cite{Freitas2004}.  The first
strategy transforms the problem into a single-objective one, usually by
weighting each objective and adding them up.  The lexicographic approach assigns
a priority to each objective and then optimizes the objectives in that order.
Hence, when comparing two candidate solutions, the highest-priority objective is
compared and, if equivalent, the second objective is compared.  If the second
objective is also equivalent between the solutions, the third
one is used, and so on.  Both the weighted-formula and the lexicographic
strategies return only one solution for the problem.  The Pareto
methods use different mathematical tools to evaluate the candidate solution,
finding a set of non-dominated solutions.  A solution
is said to be non-dominated if it is not worse than any other solution
concerning all the criteria~\cite{Freitas2004}.

Since it is not trivial the difference of scale between our objective functions,
we opted to use only a
lexicographic and a Pareto method.

We design two population-based
optimization algorithms.  Specifically, we consider the use of two
metaheuristics:
\begin{itemize}
  \item a lexicographic genetic algorithm (GA); and
 \item the improved strength Pareto evolutionary algorithm (SPEA2)~\cite{Zitzler2001}.
\end{itemize}

Although the methods are different, both approaches share many properties --
individual representation, population initialization, operators of mutation,
recombination and selection -- which are explained in \cref{sub:design}.
The main difference between them is the fitness evaluation.

In the GA, the individuals are ordered lexicographically,
that is, ordered primarily by the first objective function and, in case of tie,
the second objective.  SPEA2, however, consider not only the Pareto front but
also the density of the solutions.

\subsection{Lexicographic genetic algorithm review}
\label{sub:traditional}

A GA has the following steps~\cite{Srinivas1994}:
\begin{algorithmic}[1]
  \State $X \gets$ \Call{InitialPopulation}{{}}
  \While{\Call{StopCondition}{{}} = false}
    \State \Call{Evaluate}{$X$}
    \State $X_{\text{next}} \gets$ \Call{Elitism}{$X$}
    \While{$\card{X_{\text{next}}} < \card{X}$}
      \State parents $\gets$ \Call{Select}{$X$}
      \State children $\gets$ \Call{Recombine}{parents}
      \State \Call{Mutate}{children}
      \State $X_{\text{next}} \gets$ $X_{\text{next}} \cup$ children
    \EndWhile
    \State $X \gets X_{\text{next}}$
  \EndWhile.
\end{algorithmic}

In words, a random population of candidate solutions is generated as the first
step.  Then,
while the stop condition is not met, the next generation of individuals
comprises the best individuals of the previous generation and the individuals
originated by recombining and mutating parents selected from the previous
generation.

In the lexicographic approach, the only difference is during the evaluation of
the candidate solution~\cite{CoelloCoello2002}, where the best individuals are
decided lexicographically.

In our specific problem, a solution $\mathcal{Y}_1$ is better than
$\mathcal{Y}_2$ if
\begin{itemize}
  \item $D(\mathcal{Y}_1) > D(\mathcal{Y}_1)$; or
  \item $D(\mathcal{Y}_1) = D(\mathcal{Y}_1)$
    and $\Disproportion\big(\FSNetwork, \FSNetwork(\mathcal{Y}_1)\big) <
    \Disproportion\big(\FSNetwork, \FSNetwork(\mathcal{Y}_1)\big)$.
\end{itemize}

\subsection{Improved Strength Pareto Evolutionary Algorithm review}
\label{sub:spea2}

SPEA2 works similarly to a GA.  The major difference is that it
keeps an archive with the candidate solutions for the Pareto set.
If the number of non-dominated solutions is greater than the limit of the
archive, some solutions are discarded.  Such operation is called truncation.
The truncation operator tries to maintain the candidate solutions uniformly
distributed along the Pareto front~\cite{Zitzler2001}.

As indicated by the authors, we select individuals by employing binary
tournament.  Also, let $A$ be the archive size, we fix the parameter $k =
\sqrt{\text{A}}$ in the truncation operator~\cite{Zitzler2001}.

\subsection{Metaheuristic design}
\label{sub:design}

The common implementation characteristics of our metaheuristic is exposed in
the next items.

\subsubsection{Individual representation}

In our problem, each solution is a set $\mathcal{Y}$ of zero or more and-features.
If we enumerate every possible and-feature, the solution can also be viewed as a
binary vector with entries 1 for the present and-features.

\subsubsection{Population initialization}
\label{sub:sample}

Given $\mu, \sigma>0$, we sample, without replacement, $\lfloor M \rfloor$
random and-features to compose each candidate solution $\mathcal{Y}$ such that
$M \sim \mathcal{N}(\mu, \sigma)$.  And-features are sampled so that the
probability of having order $q\geq 2$ is
\[
  \frac{q-1}{q!}\text{.}
\]

\subsubsection{Recombination operator}

We use the uniform crossover operator with two
parents generating two children.  In \cref{sub:performance}, we
show how to implement it efficiently using our set representation.

\subsubsection{Selection operator}

The binary tournament method is chosen to select the parent that
go to the recombination step.

\subsubsection{Mutation operator}

We formulated the following specific mutation operator to exploit
the characteristics of our problem.

Given a candidate solution $\mathcal{Y}$, we apply $\eta \in \{0, 1, 2, \dots\}$
random changes in the individual.  For each change, there is a equal probability
of either
\begin{itemize}
  \item trying to add a new and-feature; or
  \item trying to remove an and-feature $v \in \mathcal{Y}$; or
  \item trying to modify an and-feature $v \in \mathcal{Y}$.
\end{itemize}

When trying to add a new feature, an and-feature $v$ is sampled and the solution is
updated to $\mathcal{Y} \cup \{v\}$.  Note that the candidate solution
$\mathcal{Y}$ may not change if the and-feature was previously present.

If trying to remove a feature, each and-feature $v \in \mathcal{Y}$ has
probability
$\left(\card{\mathcal{Y}}+1\right)^{-1}$ of being selected.  In this case, the
candidate solution is updated to $\mathcal{Y} \setminus \{v\}$.  Note that,
with probability $\left(\card{\mathcal{Y}}+1\right)^{-1}$, the individual do not
change.

Finally, in the last case, an and-feature $v(\vec{a})\in\mathcal{Y}$ with order
$q = \sum_j{a_j}$ is selected uniformly to be modified.  Once the and-feature is
selected, a modified and-feature $v'(\vec{a}')$ will be produced.  Two cases may
happen:
\begin{inparaenum}[a)]
  \item with chance $\frac{1}{q}$, an index $j'\in[1, D]$ is selected uniformly,
    and $\vec{a}'$ is
    \begin{equation*}
      \begin{dcases*}
        a'_{j'} = 1 & {} \\
        a'_j = a_j & $\forall j \neq j'$\text{;}
      \end{dcases*}
    \end{equation*}
  \item with chance $\frac{q-1}{q}$, two indexes $j' \in \{j \pipe a_j = 1\}$
    and $j'' \in [1, D]$ are chosen uniformly.  The modified and-feature is
\begin{equation*}
  \begin{dcases*}
    a'_{j''} = a_{j'} & {}\\
    a'_{j'} = a_{j''} & {}\\
    a'_j = a_j & $\forall j \not\in \{j', j''\}$
  \end{dcases*}
\end{equation*}
\end{inparaenum}

The first case will include one more term into the and-feature if $a_{j'} = 0$.
The second one swaps two elements of $\vec{a}$ and, effectively, takes effect
when $a_{j''} = 0$.
The candidate solutions is then updated to $\left(\mathcal{Y} \setminus
\{v\}\right) \cup \{v'\}$.  The size $\card{\mathcal{Y}}$ is never increased, and
the candidate solution will be preserved if $v = v'$.

\subsubsection{Performance considerations}
\label{sub:performance}

Although, we can view both solutions and and-features as binary vectors, the set
representation is more practical because of the high space-complexity of the
problem.  Moreover, there is no need to store entries for and-features that
lack connections.  Instead of just ignoring them, we exploit the evaluation step
to determine which and-features are useless and remove them from the set.

\newcommand{\Parent}[1]{\mathcal{Y}_{#1}^{\text{parent}}}
\newcommand{\Child}[1]{\mathcal{Y}_{#1}^{\text{child}}}

Also, using the set representation, the crossover of the candidate solutions
$\Parent{1}$ and $\Parent{2}$ can be
implemented efficiently with steps
\begin{algorithmic}[1]
  \State $\Child{1}, \Child{2} \gets \Parent{1} \cup \Parent{2}$
  \For{$v \in \Parent{1} \triangle \Parent{2}$}
    \If{\Call{SampleUniform}{$0$, $1$} ${} < 0.5$}
      \State $\Child{1} \gets \Child{1} \cup \{v\}$
    \Else
      \State $\Child{2} \gets \Child{2} \cup \{v\}$
    \EndIf
  \EndFor
\end{algorithmic}
where $\triangle$ stands for the symmetric difference operator.

Finally, to conform to the requirements in \cref{sub:sample},
one can sample each candidate solution $\mathcal{Y}$ efficiently with steps
\begin{algorithmic}[1]
  \State $v \gets$ \Call{SampleAndFeature}{{}}
  \State $\mathcal{Y} \gets \{v\}$
  \Loop{}
    \State $v \gets$ \Call{SampleAndFeature}{{}}
    \State $\mathcal{Y} \gets \mathcal{Y} \cup \{v\}$
    \If{\Call{SampleUniform}{$0$, $1$} ${} < 1 - \frac{1}{\card{\mathcal{Y}}}$}
      \State \textbf{break}
    \EndIf
  \EndLoop
\end{algorithmic}
where and-features are sampled without replacement.

%% file: result.tex

\section{Experimental results}
\label{sec:results}

In this section, we present applications of our feature learning technique in
the two major categories in machine learning: supervised and
unsupervised learning.

First, we illustrate the optimization process in the famous Iris dataset.  In
this example, we also conduct community detection in the original feature--sample
network obtained from the dataset and in the enhanced one.

Then, we observe
the increase of the accuracy obtained by the $k$-nearest neighbors~\cite{Bishop2006} classifier
in other four UCI datasets.

\input{example.tex}
\input{performance.tex}

%% file: example.tex

\subsection{Enhanced community detection and clustering}
\label{sub:unsup}

The UCI Iris dataset~\cite{Lichman2013} contains 150 samples that describe the sepals and
petals of individual Iris flowers.  The flowers are either \emph{Iris setosa}, \emph{Iris
virginica}, or \emph{Iris versicolor}.  In \cite{Verri2016}, we construct a
feature--sample network from this dataset by discretizing the features.

\begin{figure}
  \centering
  \includegraphics{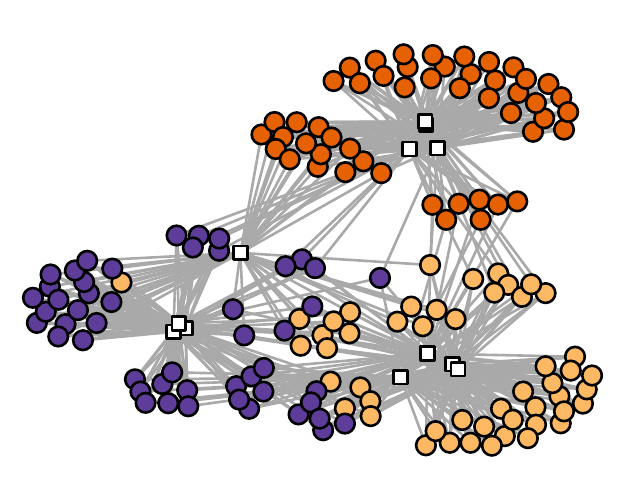}
  \caption{
    \label{fig:iris-network}
    Feature–sample network for Iris dataset.   Circles are vertices associated with
    samples and squares with features.  Colors represent the classes.
  }
\end{figure}

\Cref{fig:iris-network} shows the generated network.  Each color represents a
different class.  Circles represent samples and squares, features. We use that
same network as an input to both algorithms -- using SPEA2 and the lexicographic
GA.

\begin{figure}
  \centering
  \includegraphics{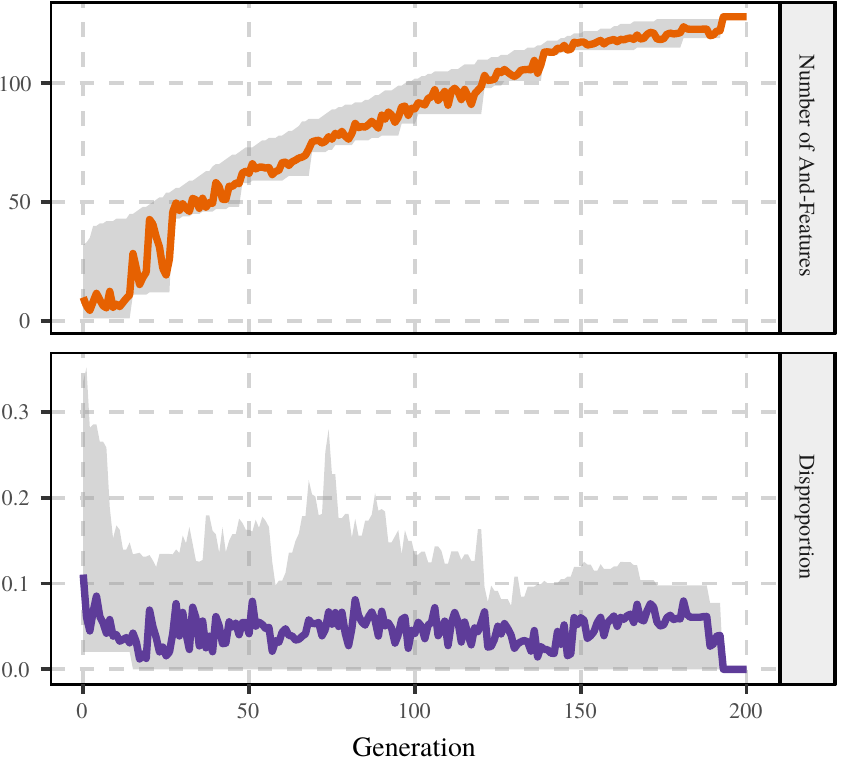}
  \caption{
    \label{fig:iris-evolution-spea2}
    Evolution of the number of and-features and disproportion along the
    generations of the SPEA2 algorithm with Iris dataset
    as the input network.  Solid lines are the average disproportion and number
    of and-features of the non-dominated solution at a given generation.
    Shadows cover the range of the measurements.
  }
\end{figure}

\begin{figure}
  \centering
  \includegraphics{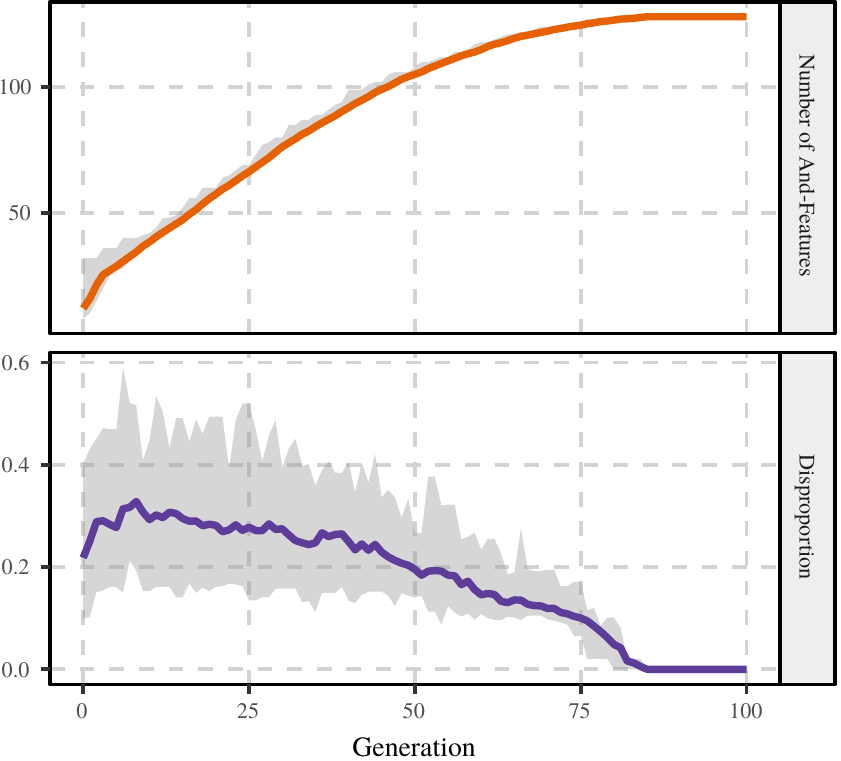}
  \caption{
    \label{fig:iris-evolution-ga}
    Evolution of the number of and-features and disproportion along the
    generations of the lexicographic GA with Iris dataset
    as the input network.  Solid lines are the average disproportion and number
    of and-features of the best $100$ solutions (elitism) at a given generation.
    Shadows cover the range of the measurements.
  }
\end{figure}

\subsubsection{Evolution of the candidate solutions}
We execute the optimization process once for each strategy.  We fix the population
in $1000$ individuals.  For SPEA2, the archive has size $100$ and, for the
lexicographic GA, we keep the $100$ best solutions over the generations.  In the
initial population, we use $\mu = 10$ and $\sigma = 5$.  The recombination rate
is $0.6$ and the apply $\eta = 1$ random changes in each generated individual.

\Cref{fig:iris-evolution-spea2,fig:iris-evolution-ga} describe the obtained
result.  Both disproportion and number of discovered and-feature are depicted
along the generations.  Solid lines are the average result in the population and
shadows cover the range -- from minimum to maximum -- of each measurement.  The
results include only the non-dominated solutions in SPEA2 and the $100$ best
solutions in the lexicographic GA.

Using both strategies, we could reach the optimal solution: $128$ and-features with at
least one connection and $0$ disproportion.  However, the optimization strategies differ
as to how to achieve this.

SPEA2 tries to find as many new and-feature while keeping the
ones with lowest values of disproportion.  When a larger set of and-features with
disproportion $0$ is discovered, such solution dominates every other solution found so
far.  Thus, we can observe ``steps'' in the evolution of the number of and-features.

In the lexicographic GA, the disproportion is only
considered when the number of discovered and-features is the same.  As a result,
the algorithm greedily produce and-features disregarding the disproportion until
it cannot find more new features to add.  It enables a faster convergence in
this case, but it may find only solutions with high disproportion when it is
unfeasible to reach the maximum number of and-feature -- which is very common in
practice.  To solve this issue in larger problems, one can set the limit in the
number of discovered features, $M_{\text{max}}$.


\begin{figure}
  \centering
  \includegraphics{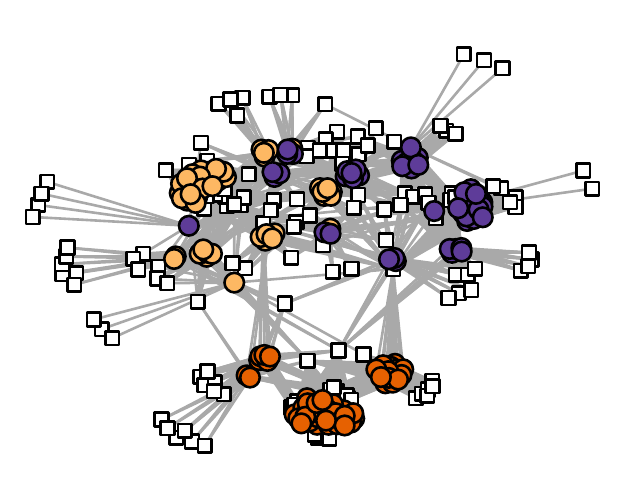}
  \caption{
    \label{fig:iris-hnetwork}
    Feature–sample network for Iris dataset with all possible 128 and-features (excluding
    those with no connections).   Circles are vertices associated with samples and squares
    with features.  Colors represent the classes.
  }
\end{figure}

The optimal enhanced feature--sample network for this dataset is in
\cref{fig:iris-hnetwork}.

\subsubsection{Community detection}


Applying a greedy community detection method~\cite{Clauset2004} in both
networks, we observe that the enhanced network has modularity $12.4\%$ ($Q =
0.561$) higher than the input network ($Q = 0.499$.)

The enhanced network can also improve clustering tasks.  If comparing the
expected class and the obtained communities, the enhanced version achieves
higher Jaccard index, $0.731$ against $0.719$.

%% file: performance.tex

\subsection{Performance enhancement in supervised learning}

\begin{table}
  \centering
  \caption{
    UCI datasets along with the number of samples $N$, features $D$, and
    possible and-feature $M$.
    \label{tab:uci-data}
  }
  \begin{tabular}{c c c r}
    \toprule
    Dataset & N & D & M \\
    \midrule
    \input{tables/uci_info.tex}
    \bottomrule
  \end{tabular}
\end{table}

We also apply our proposal in 4 classification tasks from
UCI~\cite{Lichman2013}.  \cref{tab:uci-data} presents the datasets along with
the number of samples $N$, features $D$, and possible
and-features $M$.  We highlight that it is unfeasible
to list every possible combination among the features even for small datasets.
The networks are generated as shown in \cite{Verri2016} with $3$ bins.

The optimization process is executed $15$ times for each strategy -- SPEA2 and
GA.  We fix the population size in $1000$, the archive size in $100$, and the
elitism $100$ solutions.  For the initial population, we use $\mu = 50$ and
$\sigma = 10$.  The recombination rate is $0.6$ and $\eta = 1$ mutations are
performed for each candidate solution.  We limit the number of and-feature by
$M_{max} = 100 D$.  The execution is stopped after the 1000th generation.

\begin{table}
  \centering
  \caption{
    \label{tab:variation}
    Number of and-features and disproportion obtained by the optimization
    process for both strategies.  Average and standard deviation are shown for
    each measurement.
  }
  \begin{tabular}{c c c}
    \toprule
    Dataset & LGA & SPEA2 \\
    \midrule
    \input{tables/uci_variation.tex}
    \bottomrule
  \end{tabular}
\end{table}

\Cref{tab:variation} summarizes the number of and-features and disproportion
obtained by the optimization process.  For the lexicographic GA, we show the
average and the standard deviation of the measurements among the $100$ best
individuals. For SPEA2, only the non-dominated solutions are considered.

As expected by considering the previous
study (\cref{sub:unsup},) the lexicographic GA achieved better count of
and-features  -- the maximum allowed --, but worse values of disproportion.
The candidate solutions of SPEA2 present wide variation, but consistent lower
disproportion.

\begin{table}
  \centering
  \caption{
    Number of and-features, disproportion, and strategy of the results with
    lowest disproportion and highest number of and-features.
    \label{tab:best}
  }
  \begin{tabular}{c c c}
    \toprule
    Dataset & Lowest disproportion & Highest number of and-features \\
    \midrule
    Breast 2010 & 513, 0.000 (SPEA2) & 2700, 0.129 (LGA) \\
    Ecoli       & 166, 0.011 (SPEA2) & 1900, 0.114 (LGA) \\
    Glass       & 416, 0.019 (SPEA2) & 2500, 0.130 (LGA) \\
    Wine        & 413, 0.015 (SPEA2) & 3900, 0.226 (LGA) \\
    \bottomrule
  \end{tabular}
\end{table}

\begin{table*}
  \centering
  \caption{
    Accuracy of the $k$-NN method on the UCI datasets using three different input
    networks -- original, lowest disproportion, and highest number of
    and-features.  For each setting, the best value of $k$ is shown.
    \label{tab:classification}
  }
  \begin{tabular}{c c r c r c r}
    \toprule
    Dataset & Original & ($k$) & Lowest disproportion & ($k$) & Highest number of and-features & ($k$) \\
    \midrule
    \multicolumn{7}{c}{\bf 70\% labeled} \\
    \input{tables/new_classification,p=70.tex}
    \midrule
    \multicolumn{7}{c}{\bf 80\% labeled} \\
    \input{tables/new_classification,p=80.tex}
    \bottomrule
  \end{tabular}
\end{table*}

For each one of the datasets, we take the candidate solution with highest count
of and-features and with the lowest disproportion among every solution produced.
(We ignored a few solutions with less than 100 and-features.)
Such candidate solutions, and the strategy
that has found them, are indicated in \cref{tab:best}.

We solve the classification problems for each one of the datasets using the
$k$-nearest neighbors method.  As inputs we use the interaction matrices
$\left(x_{ij} \in \{0, 1\}\right)_{ij}$ of the original network and the enhance
ones from the selected candidate solutions.  We performed $20$ split validations
with $70$ and $80\%$ of labeled samples for each case.  We also varied $k
\in \{1, 2, \dots, 20\}$.

The best results for each configuration are shown in \cref{tab:classification}.
Improvements are highlighted in bold.  Using the solution with the highest
number of and-features, we improved the classification results in all
cases.  Using less and-features but with lower
disproportion, we see improvements almost as good as using higher and-feature count.

%% file: tables/uci_info.tex
Breast 2010 & 106 & 27 &     134,217,700 \\
Ecoli & 336 & 19 &         524,268 \\
Glass & 214 & 25 &      33,554,406 \\
Wine & 178 & 39 & 549,755,813,848 \\

%% file: tables/uci_variation.tex
Reuters & $ 5991 \pm 78.7 $, $ 0.042 \pm 0.002 $\\

%% file: tables/new_classification,p=70.tex
Breast 2010 & $0.62 \pm 0.07$ & (1) & $0.63 \pm 0.07$ & (1) & $\mathbf{0.64 \pm 0.06}$ & (1)\\
Ecoli & $0.76 \pm 0.03$ & (4) & $\mathbf{0.77 \pm 0.03}$ & (4) & $0.76 \pm 0.04$ & (12)\\
Glass & $0.66 \pm 0.06$ & (7) & $0.69 \pm 0.05$ & (6) & $\mathbf{0.71 \pm 0.05}$ & (8)\\
Wine & $0.92 \pm 0.03$ & (17) & $\mathbf{0.93 \pm 0.03}$ & (3) & $\mathbf{0.93 \pm 0.02}$ & (5)\\

%% file: tables/new_classification,p=80.tex
Breast 2010 & $0.65 \pm 0.11$ & (5) & $0.62 \pm 0.10$ & (4) & $\mathbf{0.66 \pm 0.10}$ & (3)\\
Ecoli & $0.77 \pm 0.03$ & (6) & $\mathbf{0.78 \pm 0.04}$ & (4) & $\mathbf{0.78 \pm 0.04}$ & (13)\\
Glass & $0.66 \pm 0.07$ & (7) & $0.69 \pm 0.06$ & (6) & $\mathbf{0.72 \pm 0.07}$ & (9)\\
Wine & $0.93 \pm 0.05$ & (19) & $\mathbf{0.94 \pm 0.05}$ & (3) & $\mathbf{0.94 \pm 0.03}$ & (3)\\

%% file: conclusion.tex

\section{Conclusion}
\label{sec:conclusion}

In this paper, we presented an unsupervised feature learning mechanism that
works on datasets with binary features.  First, the dataset is mapped into a
feature--sample network.  Then, a multi-objective optimization process selects a set of new
vertices that correspond to new features to produce an enhanced version of the
network.

We show that the enhanced network contains more information and
can be exploited to improve the performance of
machine learning methods.

To solve the optimization problem, we opted to design population-based
metaheuristics.  We used both a lexicographic GA and the SPEA2 algorithm to find
the candidate solutions.

From the experiments, we conclude that the GA produces more new
features in fewer generations.  However, candidate solutions in SPEA2, besides
having less new features, also improved the performance of machine learning
methods.

This fact suggests that the disproportion is a good measurement of the quality
of the selected set of and-features.  In future works, we will correlate
improvement and disproportion of the solutions with the same
number of features.

Furthermore, the learning techniques used
-- fast-greedy community detection and $k$-nearest neighbors -- do not take
full advantage of the new features.  In subsequent studies, we will elaborate
learning models to exploit the enhanced feature--sample network explicitly.

\section*{Acknowledgments}

This research was supported by the São Paulo State Research Foundation (FAPESP) 
and the Brazilian National Research Council (CNPq).